\documentclass[letterpaper, 10 pt, conference]{ieeeconf}  
\IEEEoverridecommandlockouts

\usepackage[T1]{fontenc}

\usepackage{graphics} 
\usepackage{epsfig} 
\usepackage{times} 
\usepackage{amsmath} 
\usepackage{amssymb}  
\usepackage{lipsum}
\usepackage{siunitx}
\usepackage{xspace}
\usepackage{multirow}
\usepackage{graphicx}
\usepackage{makecell}
\usepackage{colortbl}
\usepackage{caption}
\usepackage{euler}
\usepackage{siunitx}
\usepackage{booktabs}
\usepackage{manfnt}
\usepackage{tikz}
\usepackage{tikzducks}
\usetikzlibrary{shapes.geometric}
\usepackage[table]{xcolor} 
\usepackage[colorlinks=true, linkcolor=customblue, urlcolor=customblue, citecolor=customblue]{hyperref}
\definecolor{customblue}{HTML}{0066CC}

\definecolor{methodorange}{HTML}{F3950D}
\newcommand{\method}{{\textbf{\texttt{HOTICE}}}\xspace}
\newcommand{\field}{HOD-PF\xspace}

\title{\LARGE \bf
HOTICE: Whole-Body Humanoid Object Transportation \\ in Cluttered Environments 
}

\author{
    \textbf{Toan Nguyen} \quad
    \textbf{Weiduo Yuan}\textsuperscript{$\dagger$} \quad
    \textbf{Siheng Zhao}\textsuperscript{$\dagger$} \quad
    \textbf{Yue Wang}\textsuperscript{$\ddagger$} \quad
    \textbf{Daniel Seita}\textsuperscript{$\ddagger$}\\[0.5em]
    $^{\dagger}$Equal Contribution \quad $^{\ddagger}$Equal Advising\\[0.5em]
    University of Southern California
}

\begin{document}


\twocolumn[{%
\renewcommand\twocolumn[1][]{#1}%
\maketitle
\begin{center}
\centering
\captionsetup{type=figure}
  \includegraphics[width=\linewidth]{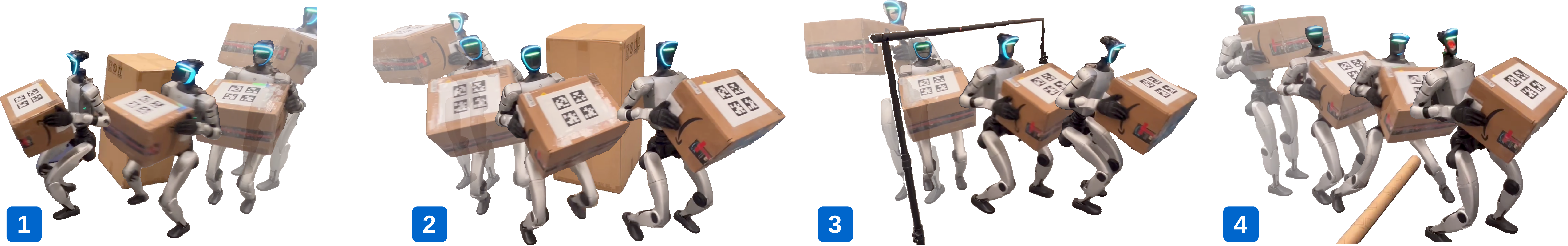}%
\end{center}%

\captionof{figure}{We introduce \textbf{\method}, a humanoid loco-manipulation method that enables humanoid robots to transport objects through cluttered environments populated by diverse types of obstacles, including side obstacles -- \textbf{(1)} and \textbf{(2)}, overhead obstacles -- \textbf{(3)}, and ground obstacles -- \textbf{(4)}. More results can be found at our anonymous website \href{https://hotice2027.github.io}{https://hotice2027.github.io}.}

\vspace{2ex}
}]

\thispagestyle{empty}
\pagestyle{empty}

\begin{abstract}
Object transportation is a fundamental capability for humanoid robots operating in real-world, human-centric environments, yet existing methods struggle when clutter constrains free space around both the robot and its carried payload. We present \method, a whole-body humanoid learning framework for transporting objects through such cluttered environments. First, we introduce Humanoid-Object Decoupled Potential Fields, which jointly encode collision-avoidance guidance for the robot and the carried object, enabling coordinated, obstacle-aware motion for both. Second, to address the large action space inherent to whole-body loco-manipulation, we design a dual-agent reinforcement learning architecture that decouples upper- and lower-body control while preserving whole-body coordination via shared state observations and rewards. To train a policy that generalizes across diverse cluttered scenes, we further employ a specialist-to-generalist distillation strategy, in which privileged teacher policies are distilled into a single deployable student policy. We evaluate \method in MuJoCo simulation and on a real Unitree G1 humanoid, demonstrating effective and robust object transportation across cluttered scenarios for objects of varying shapes. Our results show that \method reliably coordinates whole-body motion and object-aware collision avoidance, generalizing effectively to previously unseen cluttered environments while achieving strong performance in sim2real deployment.
\end{abstract}

\section{INTRODUCTION}~\label{sec:intro}
Object transportation is one of the most important problems in robotics, underpinning a wide range of applications including last-mile delivery, warehouse logistics, and household assistance. Traditionally, this problem has been mainly addressed with robotic manipulators mounted on fixed bases~\cite{brown2022amazon}, wheeled mobile platforms~\cite{mehta2025deep}, or aerial drones~\cite{lee2015path}, which achieve high reliability in structured, engineered settings. However, these platforms rely on assumptions that break down in open-world, cluttered environments: fixed-base arms are constrained by a limited workspace and cannot reposition to navigate around obstacles, wheeled bases require traversable, relatively flat ground and struggle with thresholds and narrow or irregular passages common in human-centric spaces, and drones are limited by small payload capacities and the need for open airspace.

In contrast, the humanoid form factor is well suited to these challenges~\cite{gu2026humanoid}. Whole-body loco-manipulation potentially allows humanoids to step over ground-level obstacles and adjust their posture to duck under overhead obstructions while maintaining a stable hold on the carried payload, and their human-scale footprint enables passage through narrow side clearances and dense clutter built for people. Moreover, a single humanoid embodiment can generalize across diverse environments, mitigating the need for specialized, environment-specific robot platforms.

Despite these advantages, humanoid object transportation in cluttered environments remains a challenging and under-investigated problem. Humanoid loco-manipulation requires coordinated control over both the upper and lower body, resulting in a large, high-dimensional action space that is further complicated in cluttered scenarios, where the robot must often adopt extreme, non-default poses to simultaneously maintain balance and navigate through obstacles~\cite{gu2026humanoid}. This challenge is compounded when the robot must also maintain a secure hold on the transported object, particularly when it is heavy or large, as the added dynamics and reduced clearance make safe traversal substantially more difficult.

In this work, we propose \textbf{H}umanoid \textbf{O}bject \textbf{T}ransportation \textbf{I}n \textbf{C}luttered \textbf{E}nvironments (\method), a whole-body humanoid learning framework that enables humanoid robots to transport diverse objects through complex, cluttered environments. First, we introduce the Humanoid-Object Decoupled Potential Fields (\textbf{\field}), which encode the geometric relations between the humanoid, the transported object, and surrounding obstacles. \field enables collision-free transportation by providing guidance signals for jointly avoiding humanoid–obstacle and object–obstacle collisions, while maintaining tight synchronization between the robot and the object throughout transport. Moreover, to address the large action space inherent to whole-body control, we design a dual-agent learning architecture that decouples the upper- and lower-body action spaces, substantially improving the robot's loco-manipulation capability in complex environments. Finally, we apply a scalable specialist-to-generalist distillation scheme that consolidates many scene-specific privileged teacher policies into a single generalist student policy capable of handling a wide range of obstacle configurations. Together, our system enables a Unitree G1 humanoid robot to effectively transport objects of different shapes in diverse cluttered scenarios both in MuJoCo simulation~\cite{todorov2012mujoco} and real-world deployment.

Our contributions can be summarized as follows:
\begin{itemize}
    \item We propose \field, a framework that effectively encodes the relationships between the humanoid robot, object, and obstacles for collision-free object transportation in cluttered environments.
    \item We present a dual-agent learning system to deal with the large action space difficulty of whole-body loco-manipulation under challenging conditions.
    \item By applying specialist-to-generalist distillation from various obstacle settings, our policy enables successful object transportation across various scenarios in both simulation and real-world deployment.
\end{itemize}

\section{RELATED WORK}~\label{sec:rw}
\vspace{-3ex}
\subsection{Humanoid Traversal in Cluttered Environments}
Traversing spatially constrained environments has been studied extensively in legged robotics, from model-based whole-body planning~\cite{buchanan2021perceptive,buchanan2019walking,chiu2022collision} to learned quadrupedal policies that crouch under overhangs, squeeze through gaps, and negotiate irregular voids~\cite{hoeller2024anymal,miki2024learning,xu2024dexterous,he2024agile}. Extending these capabilities to humanoids is considered harder given their higher dimensionality and narrower stability margins, yet recent work has demonstrated vision-based parkour~\cite{zhuang2024humanoid,wu2026perceptive}, contact-driven whole-body control~\cite{zhang2024wococo,yang2025omniretarget}, balance on narrow terrain \cite{wang2025beamdojo,zhao2026ladderman}, and terrain-conditioned skill composition \cite{zhang2026learning,wang2026apex}. More recently, \cite{xue2026collision} proposed HumanoidPF that encodes humanoid-obstacle relationships as collision-free motion directions, enabling object-free traversal across ground, lateral, and overhead obstacles. TANGO~\cite{li2026tango} targeted similar scenarios but for long-horizon text-driven navigation. These methods, however, assume an unencumbered robot whose collision geometry follows entirely from its own kinematics, an assumption that breaks under object transportation, where the payload extends the collision volume and perturbs balance precisely when extreme postures are required. Our \field targets this gap by decoupling the humanoid's and object's collision avoidance into separate guidance vector fields that jointly shape a collision-free carrying motion in cluttered environments.

\subsection{Humanoid Object Transportation}
Humanoid object transportation was first studied as a planning problem, in which collision-free trajectories are computed jointly for the robot and the carried object~\cite{yoshida2008planning, harada2005humanoid}. Such methods can be highly reliable, but typically rely on simplified dynamics assumptions. Learning-based approaches relax these assumptions and generalize more readily across objects and scene configurations, from sim2real box transport under varying weight and geometry~\cite{dao2024sim,kim2026humanoid} to contact-based whole-body strategies that embrace bulky objects~\cite{zheng2025embracing,bi2026genhoi} and load-aware controllers that stabilize locomotion against payload-induced disturbances~\cite{fu2026load,kang2026splitadapter}. More recent systems incorporate exteroceptive perception, enabling humanoids to carry objects over uneven terrain~\cite{cui2026pilot,zhang2026rpl}, execute long-horizon transport~\cite{gu2026refine,kim2026humanoid}, and perform visual loco-manipulation~\cite{yin2025visualmimic,wang2026vlk,xie2026grail}. Across both paradigms, however, existing work targets open or lightly obstructed scenes, where clearance can be recovered by walking around sparse, low-lying obstacles. Real-world clutter instead constrains the robot from the ground, sides, and above simultaneously, so both the robot and its carried object must fit through what free space remains. Our \method targets this regime, enabling effective humanoid object transportation through complex, highly constrained scenes.

\section{PROBLEM STATEMENT}
We investigate a humanoid loco-manipulation object transportation setting. Given a cluttered environment populated by obstacles of different types (side, ground, and overhead obstacles), the robot starts at a starting position $\mathbf{s}\in \mathbb{R}^3$ holding an object $\mathscr{O}$ and is required to transport $\mathscr{O}$ to a target position $\mathbf{g}\in \mathbb{R}^3$ without colliding with the obstacles $\mathscr{C}$. The object $\mathscr{O}$ can be of different shapes. In this paper, we mainly refer to $\mathscr{O}$ as boxes of cuboid shape, but our method, as demonstrated later in Section~\ref{sec:exps}, is highly generalizable to objects of other shapes. Note that since we mainly care about the transportation of the object and collision avoidance during the traversal, we assume the object $\mathscr{O}$ is already held by the robot at the beginning of the task in our method description and experiments. However, we do train a policy to pick up the object from a support surface to facilitate the complete pipeline of object picking and transporting to the goal position. The task is considered successful if the robot reaches the goal $\mathbf{g}$ while still holding $\mathscr{O}$, without either the robot or $\mathscr{O}$ colliding with obstacles along the way; any robot fall, drop of the object, or collision constitutes a failure.

\begin{figure*}[t]
    \centering
    \includegraphics[width=\linewidth]{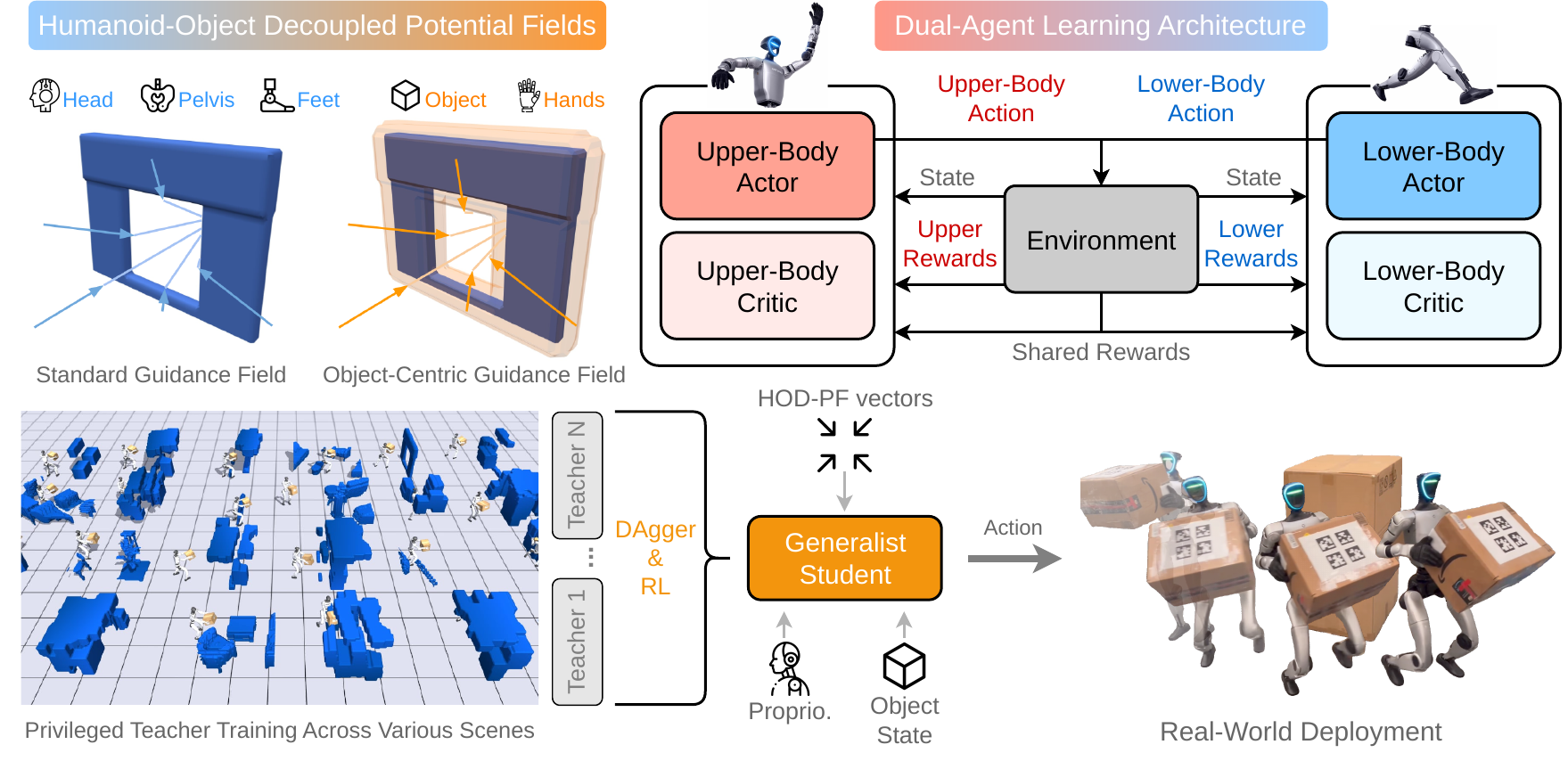}
    \caption{\textbf{Method Overview.} \method is composed of three main components. \textbf{Humanoid-Object Decoupled Potential Fields} \textit{(top left)}: a Standard Guidance Field guides the robot's head, pelvis, and feet, while an Object-Centric Guidance Field guides the carried object and the robot's hands.
    \textbf{Dual-Agent Learning Architecture} \textit{(top right)}: the whole-body control is split between an upper-body actor-critic and a lower-body actor-critic, each with its own body-specific reward set, but remain coordinated via unified state observations and shared whole-body rewards.
    \textbf{Specialist-to-Generalist Training} \textit{(bottom)}: privileged teacher policies are trained across diverse procedurally generated and realistic scenes, then distilled via DAgger and RL fine-tuning into a single generalist student policy that is deployed on the real Unitree G1 humanoid robot.
    }
    \label{fig:method}
\end{figure*}

\section{METHOD}~\label{sec:method}
We present \textbf{H}umanoid \textbf{O}bject \textbf{T}ransportation \textbf{I}n \textbf{C}luttered \textbf{E}nvironments (\method), an RL whole-body humanoid learning framework for transporting objects while avoiding collisions with diverse obstacles. \method consists of three main components: Humanoid-Object Decoupled Potential Fields (HOD-PF) that provide obstacle-aware guidance for both the robot and the carried object (Section~\ref{subsec:hod-pf}), a dual-agent learning architecture that decouples upper- and lower-body control to manage the large whole-body action space (Section~\ref{subsec:dual-agent}), and a specialist-to-generalist distillation scheme that trains scene-specific teachers and distills them into a single generalizable policy (Section~\ref{subsec:specialist-to-generalist}).
See Figure~\ref{fig:method} for an overview of our full proposed system.


\subsection{Humanoid-Object Decoupled Potential Fields}~\label{subsec:hod-pf}
Our \field builds on, and significantly extends, the Humanoid Potential Field (HumanoidPF) introduced in~\cite{xue2026collision}. Fundamentally, HumanoidPF instantiates the Artificial Potential Field~\cite{khatib1986real} by querying it at multiple body keypoints. Despite achieving strong results in collision-free humanoid traversal, HumanoidPF is designed only to control the robot's legs and does not account for a carried object. In contrast, our proposed \field is designed for whole-body control and object transportation, where the object adds considerable payload mass and volume that make the task substantially more difficult. Specifically, \field is composed of two decoupled guidance fields: a Standard Guidance Field that drives the robot's head, pelvis, and feet to enable traversal and collision avoidance, and an Object-Centric Guidance Field that guides the object's motion while maintaining robot-object synchronization during traversal.

\begin{figure}
    \centering
    \includegraphics[width=\linewidth]{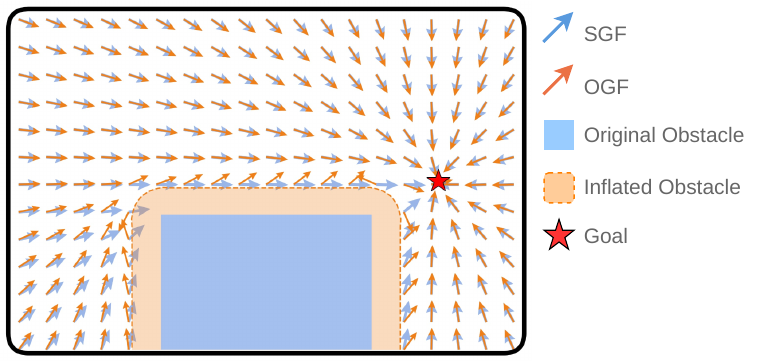}
    \caption{\textbf{Humanoid-Object Decoupled Potential Fields.} The Object-Centric Guidance Field (OGF) provides earlier collision-avoidance signals and affords the robot more headroom than the Standard Guidance Field (SGF).}
    \label{fig:fields}
\end{figure}

\textbf{Standard Guidance Field.} We follow HumanoidPF~\cite{xue2026collision} to construct our Standard Guidance Field (SGF), which is derived from an attractive field $\mathbf{U}_\text{att}: \mathbb{R}^3 \to \mathbb{R}$ and a repulsive field $\mathbf{U}_\text{rep}: \mathbb{R}^3 \to \mathbb{R}$. In particular, at any position $\mathbf{x}$ in the 3D space, the values of $\mathbf{U}_\text{att}$ and $\mathbf{U}_\text{rep}$ at $\mathbf{x}$ are computed as:
\begin{align}
    \mathbf{U}_\text{att}(\mathbf{x}) &= \eta \text{D}_\text{geo}(\mathbf{x},\mathbf{g}), \\[2mm]
    \mathbf{U}_{\text{rep}}(\mathbf{x}) &=
\begin{cases}
\dfrac{1}{2}\xi\left(\dfrac{1}{\text{sdf}(\mathbf{x})} - \dfrac{1}{d_0}\right)^2, & \text{sdf}(\mathbf{x}) \le d_0, \\[2mm]
0, & \text{sdf}(\mathbf{x}) > d_0.
\end{cases}
\end{align}
Here, $\text{D}_\text{geo}(\mathbf{x},\mathbf{g})$ is the geodesic distance between $\mathbf{x}$ and the goal $\mathbf{g}$, i.e., the length of the shortest path from $\mathbf{x}$ to $\mathbf{g}$ without intersecting obstacles $\mathscr{C}$, $\text{sdf}(\mathbf{x})$ is the signed distance function of $\mathbf{x}$ with respect to $\mathscr{C}$, $\text{d}_0$ determines the influence range of obstacles, and $\eta$ and $\xi$ are scaling factors. We set $\text{d}_0=0.2\text{m}$, $\eta=0.6$, and $\xi=1 \times 10^{-3}$, inherited from~\cite{xue2026collision}. Intuitively, $\mathbf{U}_\text{att}$ decreases toward $\mathbf{g}$, while $\mathbf{U}_\text{rep}$ grows sharply
as the robot approaches $\mathscr{C}$. The SGF $\mathbf{F}_\text{standard}: \mathbb{R}^3
\to \mathbb{R}^3$ is then obtained as the negative gradient of their sum:
\begin{equation}
    \mathbf{F}_\text{standard}(\mathbf{x}) = -\nabla_\mathbf{x} \Big( \mathbf{U}_\text{att}(\mathbf{x}) + \mathbf{U}_\text{rep}(\mathbf{x}) \Big).
\end{equation}
At each position $\mathbf{x}$, $\mathbf{F}_\text{standard}(\mathbf{x})$ therefore points in the direction of steepest decrease of the combined potential, pulling the robot toward
$\mathbf{g}$ while pushing it away from obstacles $\mathscr{C}$.
An example visualization of $\mathbf{F}_\text{standard}$ can be seen in Figure~\ref{fig:fields}. Our RL policy's input then contains the $\mathbf{F}_\text{standard}$ guidance vectors queried at the robot's head, shoulders, torso, root (pelvis), knees, and feet. In addition, the policy is optimized with rewards that encourage each body part to move along the direction indicated by its corresponding guidance vector. Concretely, let $\mathbf{x}$ be the 3D position of a robot body part among head, pelvis, and feet, and let $\mathbf{v}$ be that part's Cartesian velocity, the policy is trained to maximize the cosine similarity between $\mathbf{F}_\text{standard}(\mathbf{x})$ and $\mathbf{v}$.

\textbf{Object-Centric Guidance Field.} In addition to SGF, we implement a second vector field, the Object-Centric Guidance Field (OGF), which provides collision-avoidance signals specifically for the carried object $\mathscr{O}$ and the robot's hands. To construct OGF, we first inflate the obstacles $\mathscr{C}$ by a margin $m$ to obtain $\mathscr{C}_\text{inflated}$. The attractive field, repulsive field, and the resulting $\mathbf{F}_\text{object}$ guidance field are then computed in the same manner as in SGF, but with respect to $\mathscr{C}_\text{inflated}$. Figure~\ref{fig:fields} shows an example comparison between the resulting SGF and OGF. Computing OGF with respect to $\mathscr{C}_\text{inflated}$ offers two main benefits. First, because the object is not directly actuated by the robot's joints, the margin adds robustness to real-world conditions such as sensor noise and control delay. Second, it makes the object's avoidance behavior more anticipatory: unlike other robot parts, which can react just before contact, e.g., feet stepping over a ground bar only as it is reached, the robot must begin repositioning the object earlier, since a heavy or large object is considerably harder to maneuver once close to an obstacle. We set $m=8\text{cm}$ as it balances between collision avoidance performance and traversal feasibility.

On the carried object $\mathscr{O}$, we query $\mathbf{F}_\text{object}$ guidance vectors at its keypoints, e.g., the eight corners of box objects. Note that our state-based RL policy is lightweight (MLP-based), thus supporting a wide range of keypoint counts without hurting latency and enabling controllable granularity. In addition to object keypoints, $\mathbf{F}_\text{object}$ vectors are also queried for the robot's hands that control the object to maintain robust robot-object interaction and synchronization. Those guidance vectors are included in the policy's input along with $\mathbf{F}_\text{standard}$ vectors of other body parts. OGF-based steering reward functions are also implemented. Concretely, let $\mathbf{x}$ and $\mathbf{v}$ be the position and velocity of a robot hand or an object keypoint, the policy is trained to maximize the cosine similarity between $\left [\mathbf{F}_\text{object}(\mathbf{x}) - \mathbf{c} \right ]$ and $\left [\mathbf{v} - \mathbf{c} \right ]$, where $\mathbf{c}$ is the overall navigation command aggregated from $\mathbf{F}_\text{standard}$ vectors of the robot's head, pelvis, and feet. Subtracting $\mathbf{c}$ removes the motion that locomotion already produces, so the reward credits only the residual, obstacle-driven adjustments the hands are actually responsible for.

\subsection{Dual-Agent Learning Architecture}~\label{subsec:dual-agent}
Although the proposed HOD-PF provides helpful and specialized collision-avoidance signals, simultaneously traversing complex scene geometries and maintaining control over an unwieldy object remains a significant challenge for whole-body loco-manipulation learning. To tackle this, inspired by~\cite{falcon}, instead of treating the whole RL policy as a single learning agent, \method employs a dual-agent learning architecture, splitting the high-dimensional action space into an upper-body part that controls 8 arm joints and a lower-body part that controls 12 leg joints. Note that we exclude the waist and wrist joints, which are found to be unnecessary for our setting. In our dual-agent design, each agent maintains its own actor and privileged critic, but both read the same observation and share termination conditions, so the two policies stay coupled through the environment rather than through a monolithic network. The two actors emit $\mathbf{a}_{\text{low}} \in \mathbb{R}^{12}$ and $\mathbf{a}_{\text{up}} \in \mathbb{R}^{8}$, concatenated into the full 20D action $\mathbf{a} = [\mathbf{a}_{\text{low}}, \mathbf{a}_{\text{up}}]$, which is then fed into a PD controller to produce torque commands. We partition the reward terms into a lower set $\mathcal{L}$ (locomotion, balance), an upper set $\mathcal{U}$ (object reaching, grasp maintenance), and a shared set $\mathcal{S}$ (object lifting and \field rewards, whose values depend jointly on arm articulation and torso pose), yielding two separate reward streams:
\begin{equation}
r_{\text{low}} = \sum_{i \in \mathcal{L} \cup \mathcal{S}} w_i r_i, \qquad r_{\text{up}} = \sum_{i \in \mathcal{U} \cup \mathcal{S}} w_i r_i .
\end{equation}
Each agent then forms its own advantage estimate (denoted as $\hat{A}_{\text{low}}$ and $\hat{A}_{\text{up}}$) from its own reward stream and value function, and the two PPO~\cite{schulman2017proximal} clipped surrogate objectives are optimized jointly:
\begin{equation}
\mathcal{J} = \mathcal{J}^{\text{PPO}}_{\text{low}}\big(\hat{A}_{\text{low}}\big) + \mathcal{J}^{\text{PPO}}_{\text{up}}\big(\hat{A}_{\text{up}}\big).
\end{equation}
Because the shared terms enter both reward streams, the two agents remain jointly accountable for the carry, while the locomotion and grasping signals no longer compete inside a single advantage estimate.

\begin{figure*}[t]
    \includegraphics[width=\linewidth]{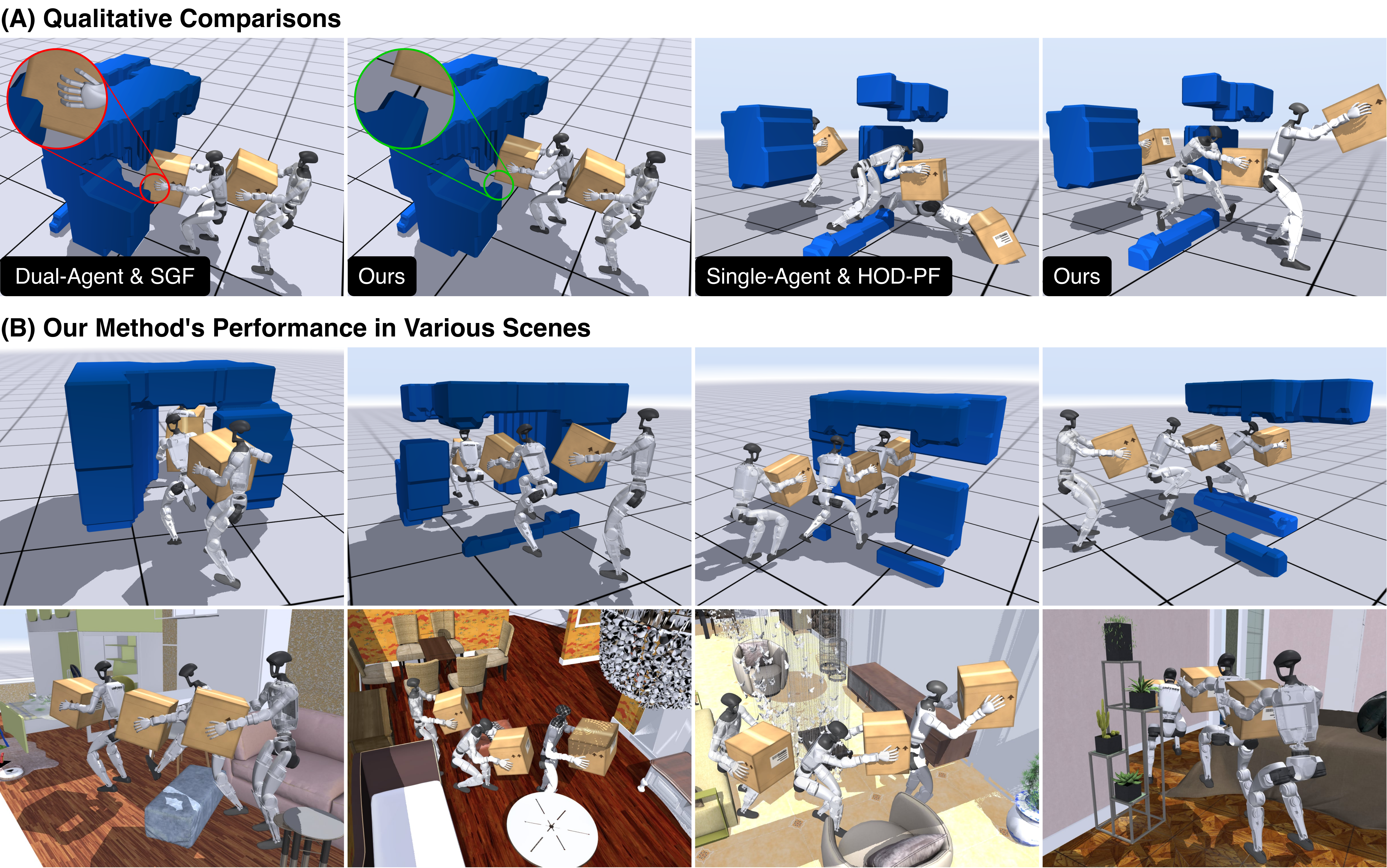}
    \caption{\textbf{Qualitative Results in Simulation.} \textbf{(A)}-\textit{left comparison} -- The proposed decoupled potential fields enable \method to more reliably avoid collisions during object transportation. \textbf{(A)}-\textit{right comparison} -- Our method is capable of maintaining the robot's balance and stability under challenging obstacle conditions thanks to the dual-agent architecture. \textbf{(B)} -- Our method effectively handles a wide variety of challenging procedurally generated scenes \textit{(upper row)} and realistic indoor scenes \textit{(lower row)}, without any scene-specific reward tuning.}
    \label{fig:qualitative_results}
\end{figure*}

\subsection{Specialist-to-Generalist Training}~\label{subsec:specialist-to-generalist}
To enable a policy that can effectively handle a wide range of cluttered settings, we apply a scalable specialist-to-generalist distillation similar to~\cite{xue2026collision}. More specifically, we first train in parallel multiple scene-specific teacher policies that have access to noiseless, non-delayed privileged states. We employ the scene construction pipeline in~\cite{xue2026collision}, which supports procedurally generated scenes and realistic scenes cropped from the 3D-FRONT dataset~\cite{fu20213d}. While the former enables the creation of highly complex scenes that push the limits of the learned policy, the latter improves generalization to real-world scenarios. Note that we do not perform any scene-specific reward tuning in training teacher policies, as our decoupled potential fields already serve as a generalizable representation and reward structure. The trained teacher policies are then distilled into a single deployable generalist student policy via DAgger distillation~\cite{ross2011reduction}. On top of DAgger distillation, we also do PPO training to further refine the distilled student policy. The PPO training of the student policy follows the same dual-agent framework used to
train its teachers, described in Section~\ref{subsec:dual-agent}. To mitigate the sim2real gap for the student policy, we simulate sensor noises, control delays, etc., in its training. We employ MuJoCo Playground~\cite{zakka2025mujoco} for our policy training.


\section{EXPERIMENTS}~\label{sec:exps}
We conduct extensive experiments in both simulation (Section~\ref{subsec:sim_exps}) and the real world (Section~\ref{subsec:real}) to validate the performance and generalizability of our proposed \method.

\subsection{Simulation Experiments}~\label{subsec:sim_exps}
\textbf{Main Experiments.} We first investigate the effectiveness of our method in enabling humanoid robots to transport objects in cluttered scenes while avoiding collisions with complex obstacles. Since no prior method directly addresses our problem setting, we assess the significance of our key contributions through ablations of \method. In particular, we design three baselines to compare with our method: (1) a variant that employs only the Standard Guidance Field, effectively a naive application of HumanoidPF~\cite{xue2026collision}, to all keypoints of the robot and the carried object, without a dedicated Object-Centric Guidance Field; (2) a single-agent variant that uses \field but replaces our dual-agent architecture with a unified policy; and (3) a variant using neither the OGF nor the dual-agent architecture.

For each of the four methods, we train teacher policies for 400M timesteps on 20 scenes of diverse types of obstacles and geometries and then distill them into a student policy using our distillation process described in Section~\ref{subsec:specialist-to-generalist}. Refer to Figure~\ref{fig:qualitative_results} for some example scenes we use in this experiment.  We use cuboid-shaped boxes as our primary objects unless stated otherwise. Box mass and size are randomized, capped at a maximum mass of 3.5\,kg and a maximum size of $40 \times 40 \times 40\,\text{cm}^3$, chosen to match the payload and scale of the G1 robot and to ensure traversal feasibility under extreme conditions in our simulation scenes. In evaluation, we apply input noises, delays, and perturbations to reflect the real-world conditions. We benchmark all methods across 20 scenes, each with 1{,}000 rollouts, using two metrics: (1) success rate (SR), the percentage of trials in which the robot reaches within \SI{0.1}{\meter} of the target location within \SI{5}{\second} without falling, dropping the box, or colliding with any obstacle (either the robot or the box); and (2) mean distance to goal (mDist, \si{\meter}), the average distance between the robot's root and the target at the end of every trial.

Table~\ref{tab:quantitative_results} shows that our proposed \method outperforms other baselines on both metrics by a large margin, demonstrating the importance of both the dual-agent architecture and the Humanoid-Object Decoupled Potential Fields. More qualitative results of \method and comparisons to its ablations can be seen in Figure~\ref{fig:qualitative_results}, where the effectiveness of our method in avoiding collisions and handling diverse challenging scenarios is further illustrated.



\textbf{Generalization to Unseen Scenes.}
We distill a generalist policy from 75 specialist teacher policies trained on 75 scenes covering diverse obstacle configurations and evaluate it on 50 unseen environments. To better interpret the results, we measure the similarity between each unseen scene and the training scenes in terms of clutter configuration. Specifically, for a pair of an unseen scene and a training scene, we discretize their SGFs and OGFs at a resolution of $(\SI{0.04}{\meter})^3$ and compute the average cosine similarity between guidance vectors at corresponding positions, across both fields and all positions. The similarity between an unseen scene and the training set is then defined as the maximum similarity across all training scenes. We present SR and mDist against this similarity measure for every unseen scene.

\begin{figure}
    \includegraphics[width=\linewidth]{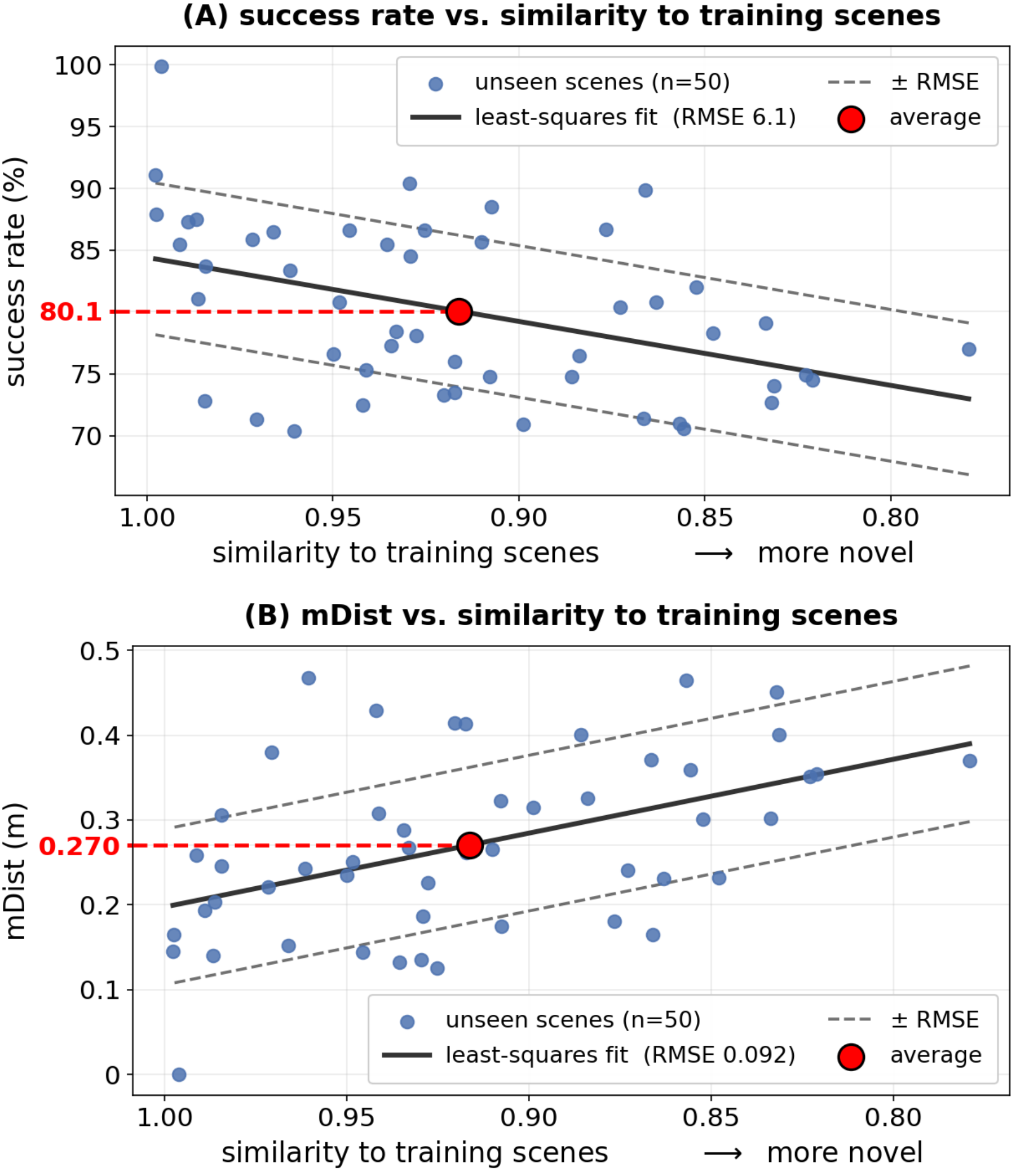}
    \caption{\textbf{Results on Unseen Scenes.} We evaluate our generalist policy on 50 unseen scenes in simulation and plot \textbf{(A)} success rate and \textbf{(B)} mean distance to goal against similarity to the training set. Least-squares fits (shaded by $\pm$RMSE) indicate a graceful degradation in both metrics with increasing novelty, while our method still maintains strong average performance.}
    \label{fig:unseen_results}
\end{figure}

The results in Figure~\ref{fig:unseen_results} indicate that success rate tends to decrease and the mean distance to goal tends to increase as scenes become less similar to the training scenes, as expected. Nonetheless, our method maintains strong performance across unseen scenes, achieving an average success rate of 80.1\% (with a floor of approximately 70\%) and an average mDist of \SI{0.27}{\meter}. Overall, these results highlight both the generalizability of \field and the effectiveness of teacher-to-student policy distillation.

\begin{table}[t]
\centering
\renewcommand{\arraystretch}{1.1}
\resizebox{\linewidth}{!}{%
\begin{tabular}{lcc}
\toprule
\textbf{Method} & \textbf{SR} (\%) $\uparrow$ & \textbf{mDist} (\si{\meter}) $\downarrow$ \\
\midrule
Single-Agent \& SGF & 58.7 & 0.296 \\
Single-Agent \& HOD-PF & 65.7 & 0.267 \\
Dual-Agent \& SGF & 77.1 & 0.195 \\
Dual-Agent \& HOD-PF (Ours) & \bf{88.5} & \bf{0.130} \\
\bottomrule
\end{tabular}
}
\caption{\textbf{Quantitative Simulation Results.} Success rates (SR) and mean distances to goal (mDist) are reported over 20{,}000 trials in 20 scenes with diverse obstacle conditions.}
\label{tab:quantitative_results}
\end{table}


\textbf{Generalization to Different Object Shapes.}
We next investigate whether \method generalizes to object shapes beyond the cuboid box, namely cylinders and spheres. The main difference between the policies for these shapes and the box policy lies in the keypoints at which the Object-Centric Guidance Field vectors are queried on the object. Specifically, for cylinder objects, we query the field at four evenly spaced points on the top rim and four on the bottom rim; for sphere objects, we use eight fixed world-frame octant directions at the sphere's radius. Since every shape uses exactly eight keypoints, the observation layout, network architecture, and action space remain unchanged across all three shapes. Beyond this, we apply only minimal shape-specific reward adaptations, e.g., for hand placement and object uprightness, while all other reward terms, weights, and hyperparameters are inherited unchanged from the box policy. We train and evaluate shape-specific policies on the same 20 scenes as in the main experiment.

\begin{figure}
    \includegraphics[width=\linewidth]{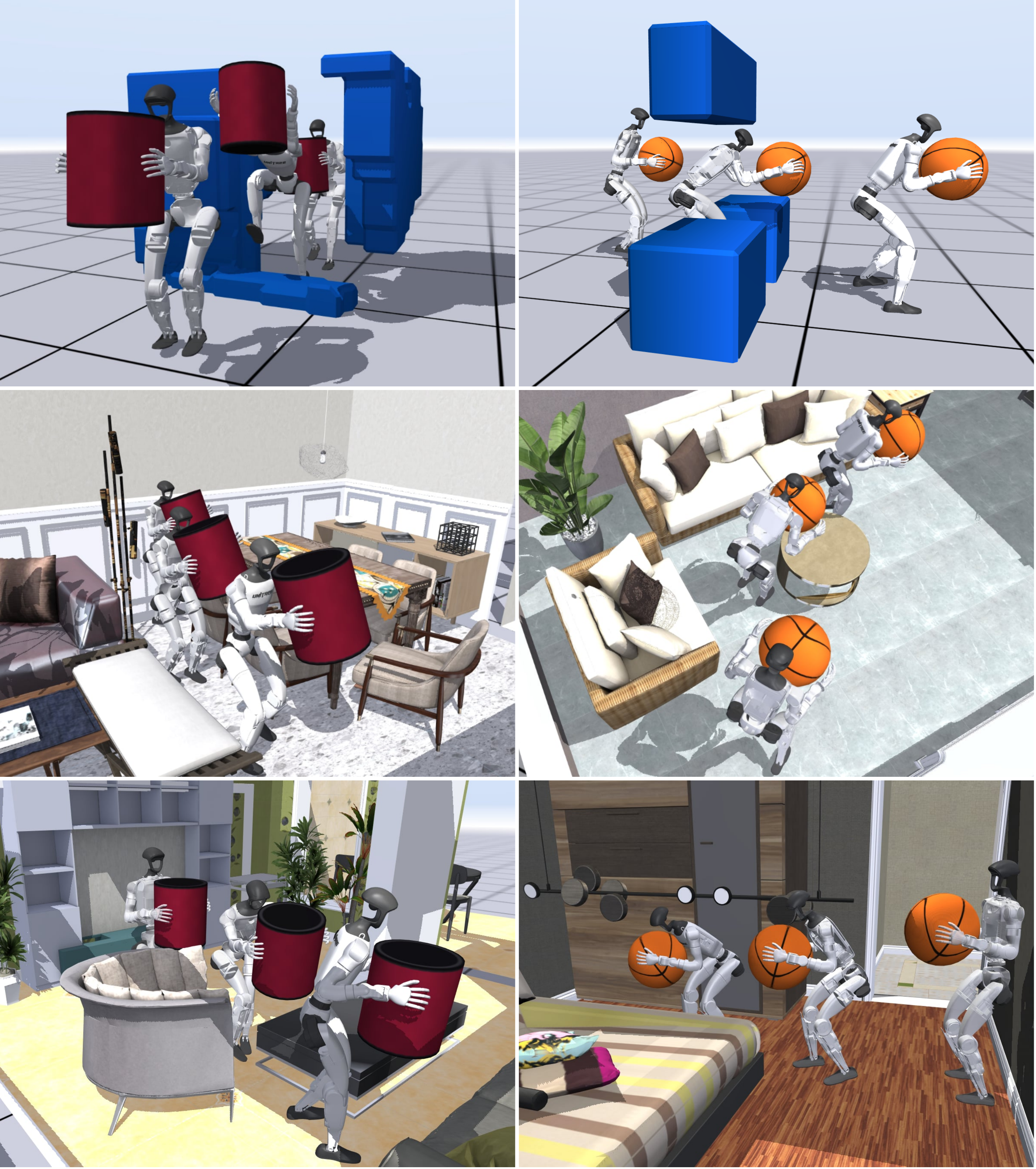}
    \caption{\textbf{Results on Other Objects.} \method is also generalizable to cylinder \textit{(left)} and sphere object shapes \textit{(right)} with only minimal shape-specific reward tuning.}
    \label{fig:more_results_cylinder_sphere}
\end{figure}

As shown in Table~\ref{tab:object_shapes}, performance on both metrics remains consistent across all three object types, demonstrating the effectiveness of our method across widely-used shape primitives. More qualitative results of the cylinder and sphere policies performing in different procedurally generated scenes and realistic scenes are illustrated in Figure~\ref{fig:more_results_cylinder_sphere}.

\begin{table}[t]
\centering
\renewcommand{\arraystretch}{1.2}
\resizebox{\linewidth}{!}{%
\begin{tabular}{lccc}
\toprule
& Cuboid \scalebox{1.4}{\mancube} & Cylinder \tikz \node[cylinder, draw, line width=0.6pt, shape border rotate=90, aspect=0.4, minimum height=0.1cm, minimum width=0.1cm] {}; & Sphere \tikz \shade[ball color=white, line width=0.6pt] (0,0) circle (0.15cm); \\
\midrule
\textbf{SR} (\%) $\uparrow$ & 88.5 & 86.7 & 87.2 \\
\textbf{mDist} (\si{\meter}) $\downarrow$ & 0.130 & 0.127 & 0.131 \\
\bottomrule
\end{tabular}
}
\caption{\textbf{Results on Different Object Shapes.} Results are reported for three shape-specific policies distilled from corresponding teachers trained on the same 20 scenes. The cuboid results match those of our method in Table~\ref{tab:quantitative_results} since we use the same policy here.}
\label{tab:object_shapes}
\end{table}

\subsection{Real-World Results}~\label{subsec:real}
\textbf{Deployment Details.}~We perform sim2real deployment for our method on a Unitree G1 robot. Obstacle scenes are first reconstructed via SLAM~\cite{xu2022fast} by walking the robot around the environment without the box. During inference, HOD-PF guidance vectors are inferred from the robot's real-time localization and joint angles, and the box pose, which we estimate via the AprilTags~\cite{olson2011apriltag} attached to the box and the robot's head camera. Note that our method could in principle support live LiDAR-based scene reconstruction, similar to~\cite{xue2026collision}. However, the G1 is equipped with only a single LiDAR sensor, whose field of view is significantly occluded by the carried box. We attribute this to a hardware limitation rather than a limitation of the method itself.

\begin{figure}
    \includegraphics[width=\linewidth]{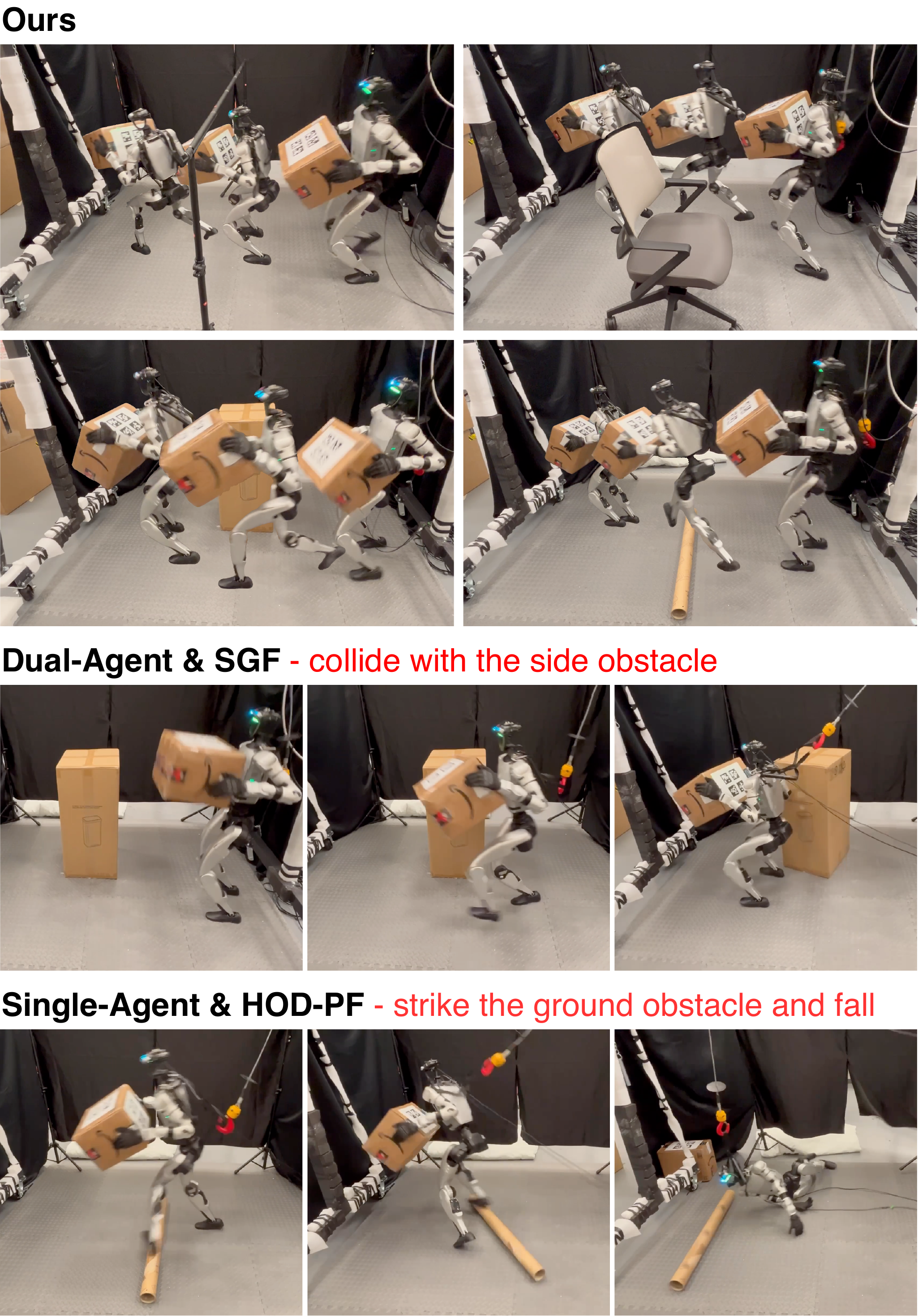}
    \caption{\textbf{Real-World Qualitative Results.} \method performs effectively across different types of obstacles, whereas the SGF-only policy is more prone to collisions, and the single-agent policy underperforms on challenging motions such as hurdling over the ground obstacle.}
    \label{fig:real_rollouts}
\end{figure}

\textbf{Real-World Experiments.}~We benchmark our method and its ablations on three representative obstacle types: side, overhead, and ground obstacles. For each obstacle type, we perform 6 trials per method. At the start of each trial, a shared box-pickup policy first picks up the box from a support surface, followed by the transportation policy; both run at \SI{50}{\hertz}. The success rate results are reported in Table~\ref{tab:real_results}. In particular, our method obtains the best real-world results, succeeding in all side obstacle trials and achieving 5/6 success rates on overhead and ground obstacle scenes. Examples of real-world rollouts are shown in Figure~\ref{fig:real_rollouts}, where \method demonstrates high performance across diverse settings, while the SGF-only policy collides more frequently and the single-agent policy struggles to remain stable during challenging motions such as stepping over the ground pipe obstacle. Overall, these results demonstrate the effectiveness of our proposed \method in real-world environments.





\begin{table}[t]
\centering
\renewcommand{\arraystretch}{1.1}
\resizebox{\linewidth}{!}{%
\begin{tabular}{lccc}
\toprule
\textbf{Method} & \textbf{Side} & \textbf{Overhead} & \textbf{Ground} \\
\midrule
Single-Agent \& SGF & 3/6 & 3/6 & 1/6 \\
Single-Agent \& HOD-PF & 6/6 & 3/6 & 2/6 \\
Dual-Agent \& SGF & 4/6 & 4/6 & 4/6 \\
Dual-Agent \& HOD-PF (Ours) & \bf{6/6} & \bf{5/6} & \bf{5/6} \\
\bottomrule
\end{tabular}
}
\caption{\textbf{Real-World Experimental Results.} For each method, we conduct 6 trials per obstacle type.}
\label{tab:real_results}
\end{table}

\section{DISCUSSION AND LIMITATIONS}~\label{sec:discussion}
Despite achieving strong performance across several settings, \method is not without limitations. First, while our method exhibits strong collision-avoidance capability, we do not model post-collision scenarios. As a result, if a collision applies a disturbance large enough to exceed the robot's recovery capacity, the robot falls. Moreover, although our loco-manipulation policy is trained to manipulate the carried object to avoid obstacles, more complex manipulation strategies remain out of reach for our method, such as carrying the object on the robot's shoulder or tucking it against the hip. We also do not tackle dynamic scenes with moving obstacles in this work. Enabling object transportation in such dynamic settings could be significantly beneficial for human-centric, open-world environments. Finally, one could investigate scenarios involving ``acceptable'' collisions alongside harder ones~\cite{ling2025impact}, for instance, the robot leaning against a wall to navigate a tight space, or brushing against a soft obstacle such as a bean bag chair without penalty. More broadly, the robot could also interact with obstacles to actively make way, e.g., using its feet to push small obstacles aside or using its back to push a door. We leave these interesting research ideas for future work.



\section{CONCLUSION}~\label{sec:conclusion}
We present \method, a whole-body humanoid learning framework for transporting objects through cluttered environments that constrain the robot and its payload from the ground, sides, and above. By encoding humanoid-obstacle and object-obstacle geometry into decoupled guidance fields, splitting whole-body control across a dual-agent architecture, and consolidating scene-specific teachers into a single generalist policy, \method achieves reliable, collision-aware transport across diverse clutter and object shapes. Extensive experiments in MuJoCo and on a Unitree G1 demonstrate strong performance on previously unseen scenes and robust sim2real transfer. We hope \method inspires follow-up work on humanoid loco-manipulation in clutter.

\section{ACKNOWLEDGEMENTS}~\label{sec:acknowledgements}
We thank Sicheng He for fruitful discussions, and Hongyi Jing and Yiqi Zhao for their help with hardware. Daniel Seita acknowledges generous support from Samsung Research America and from NSF grant \#2447397. The USC Physical Superintelligence Lab acknowledges generous supports from Toyota Research Institute, Bosch, Dolby, Google DeepMind, Capital One, Nvidia, Qualcomm and from NSF grant \#2434460. Yue Wang is also supported by a Powell Research Award.





\bibliographystyle{IEEEtran}
\bibliography{reference}

\end{document}